\documentclass[conference]{IEEEtran}

\usepackage{polyglossia}
\setmainlanguage{english} % Continue using english for rest of the document
\setotherlanguages{chinese} % Now can use \texthindi
\newfontfamily\chinesefont[
Path = fonts/,
Extension = .ttf,
UprightFont = *-Regular
]{NotoSerifSC}

\usepackage{amsmath,amssymb,amsfonts}
\usepackage{algorithmic}
\usepackage{graphicx}
\usepackage{multirow}
\usepackage{textcomp}
\usepackage{tabularx}
\usepackage{xcolor}
\usepackage{multirow}
\usepackage{makecell}
\usepackage{booktabs}
\usepackage{tcolorbox}
\usepackage{url}
\usepackage{eso-pic}

\AddToShipoutPictureBG*{%
  \AtPageLowerLeft{%
    \hspace{1cm}%
    \raisebox{1cm}[0pt][0pt]{%
        \parbox{\dimexpr\paperwidth-2cm\relax}{%{
        \footnotesize
        Copyright \textcopyright\ 2026 IEEE.  Personal use of this material is permitted.  Permission from IEEE must be obtained for all other uses, in any current or future media, including reprinting/republishing this material for advertising or promotional purposes, creating new collective works, for resale or redistribution to servers or lists, or reuse of any copyrighted component of this work in other works.
      }%
    }%
  }%
}

\tcbuselibrary{breakable} % Allows long prompts to split across page breaks smoothly

\def\BibTeX{{\rm B\kern-.05em{\sc i\kern-.025em b}\kern-.08em
    T\kern-.1667em\lower.7ex\hbox{E}\kern-.125emX}}
\begin{document}

\title{Pretrained ASR Pseudo-labeling for Noisy Police Audio\thanks{© 2026 IEEE.  Personal use of this material is permitted.  Permission from IEEE must be obtained for all other uses, in any current or future media, including reprinting/republishing this material for advertising or promotional purposes, creating new collective works, for resale or redistribution to servers or lists, or reuse of any copyrighted component of this work in other works.}\\
}

\author{
\IEEEauthorblockN{Kaavya Chaparala\IEEEauthorrefmark{1},
Su Huang\IEEEauthorrefmark{1},
Stephen L. Morgan\IEEEauthorrefmark{1},
Rhiannon N. Miller\IEEEauthorrefmark{2},
Anjalie Field\IEEEauthorrefmark{1}}
\IEEEauthorblockA{\IEEEauthorrefmark{1}\textit{Johns Hopkins University}, Baltimore, USA  \quad
\IEEEauthorrefmark{2}\textit{Providence College}, Providence, USA}
}
% \author{\IEEEauthorblockN{Kaavya Chaparala}
% \IEEEauthorblockA{\textit{Computer Science} \\
% \textit{Johns Hopkins University}\\
% Baltimore, USA \\
% kchapar1@jh.edu}
% \and
% \IEEEauthorblockN{Su Huang}
% \IEEEauthorblockA{\textit{Electrical and Computer Engineering} \\
% \textit{Johns Hopkins University}\\
% Baltimore, USA \\
% shuan148@jh.edu}
% \and
% \IEEEauthorblockN{Stephen L. Morgan}
% \IEEEauthorblockA{\textit{Sociology} \\
% \textit{Johns Hopkins University}\\
% Baltimore, USA \\
% stephen.morgan@jhu.edu}
% \and
% \IEEEauthorblockN{Rhiannon N. Miller}
% \IEEEauthorblockA{\textit{Sociology and Anthropology} \\
% \textit{Providence College}\\
% Providence, USA \\
% rmiller4@providence.edu}
% \and
% \IEEEauthorblockN{Anjalie Field}
% \IEEEauthorblockA{\textit{Computer Science} \\
% \textit{Johns Hopkins University}\\
% Baltimore, Maryland \\
% anjalief@jhu.edu}
% }

\maketitle

\begin{abstract}
Pretrained ASR systems perform poorly on noisy Broadcast Police Communication (BPC), hindering efforts to understand police decision-making. Pseudo-labeling offers an unsupervised path to improve ASR without expensive human labels, but the efficacy of this approach on very noisy domains is not known. In this work, we systematically assess the opportunities and limits of pseudo-labeling to adapt foundation ASR models (Whisper and Qwen3-ASR) to noisy BPC domain corpora from Baltimore and Chicago. We demonstrate that existing internal confidence metrics (log-probabilities and STAR scores) fail to distinguish between high and low quality BPC pseudo-labels, and we introduce an external LLM-as-a-judge filtering paradigm that leverages parametric knowledge to discard contextually implausible transcripts. Our LLM-judging filters more aggressively than internal metrics and significantly reduces WER of the pseudo-labeled training sets across the Baltimore and Chicago BPC corpora, though a substantial gap remains relative to an oracle filter. We also introduce a new cross-model pseudo-labeling paradigm where one model is finetuned with pseudo-labels from the other, and we identify this method as a promising direction for future pseudo-labeling work. 

\end{abstract}


\section{Introduction}
To address demands for transparency in law enforcement, we must acquire a deeper understanding of police activity and decision-making. Broadcast Police Communication (BPC) captures decisions as they unfold: it is used by dispatchers to direct police units, and by officers to relay information from the field \cite{morgan2025police, narayanan2024race}. However, extracting insights from this domain is not trivial. BPC is characterized
by noise, short utterances, and domain- and location-specific
terminology. While ASR for BPC has not been explored extensively, initial work suggests off-the-shelf (OTS) pretrained
ASR models like Whisper yield Word Error Rates (WER) up to 22 times higher on BPC data than on standard clean corpora like LibriSpeech \cite{panayotov2015librispeech, radford2022robustspeechrecognitionlargescale,srivastava2024speech},
making automated analysis of BPC semantic content
unreliable.  
% Units receive directions over a main  channel, and alternate channels are used for emergency situations, communication with other units, or any situation when the primary channel is unusable. 
% We speculate that some channels may even contain casual conversation between individual officers.
Manual analysis of BPC is infeasible because the data have enormous volume: 3 years of data from 1 city consists of over 91M files \cite{morgan2025police}. At the same time, labeled BPC audio is expensive to obtain for supervised finetuning, especially for every city of interest. Instead, we look to  leverage \textit{unlabeled} BPC domain data to improve ASR and enable future analysis of the semantic content.

% However, extracting insights from this data remains difficult. BPC data have enormous volume: 3 years of data from 1 city consists of over 91M files and 270K hours of audio \cite{morgan2025police}, making manual analysis infeasible.
% Furthermore, BPC is characterized by noise, short utterances, and domain- and location-specific terminology. While ASR in this domain has not been well-studied, initial work suggests off-the-shelf (OTS) pretrained ASR models like Whisper obtain  roughly 22 times higher WER on BPC than on LibriSpeech \note{cite librispeech}\cite{srivastava2024speech}, making automated at-scale analysis of BPC semantic content unreliable \cite{narayanan2024race}. While supervised finetuning does improve performance \cite{srivastava2024speech}, manual BPC transcripts are expensive to obtain for every city of interest. 

 In this work, we develop and evaluate pipelines to improve BPC ASR systems through \textit{pseudo-labeling}.
In ASR, pseudo-labeling involves taking unlabeled audio, using an ASR model to generate a transcript for it, and treating the model-generated transcript as the data label. 
The pseudo-labeled dataset is then used for training or fine-tuning.
Pseudo-labeling is a compelling approach to adapt ASR systems to the BPC domain because \textit{unlabeled} BPC data is highly accessible via unencrypted police radio channels or archival websites \cite{ahmad2023datastore},\footnote{\url{https://broadcastify.com}}
and pseudo-labeling can improve ASR across standard benchmarks \cite{kahn2020self, chen2020semi, higuchi2021momentum,manohar2021kaizen, park2020improved, xu2020iterative, Likhomanenko2020slimIPLLI}.

% (steps 1 and 3 in \ref{fig:my_single_column_figure}). 

However, the majority of the relevant pseudo-labeling studies occurred before the era of massive pretrained ASR models, and they leverage initial pseudo-labels of substantially better quality than those attainable on BPC.
% We study pseudo-labeling as a method of adapting off-the-shelf foundation ASR models to the noisy BPC domain.  
% To answer, we  pseudo-label with pretrained Whisper and Qwen3-ASR ASR models \cite{radford2022robustspeechrecognitionlargescale, shi2026Qwen3-ASR3} on BPC data from two U.S. cities: Baltimore, Maryland and Chicago, Illinois. 
Furthermore, success in pseudo-labeling is contingent on the ability to discard poor pseudo-labels \cite{kahn2020self}. Existing methods use internal model confidence metrics to assess pseudo-label quality, such as log-probabilities \cite{kahn2020self} or scores derived from attention weights \cite{hu2024selftaughtrecognizerunsupervisedadaptation}. While these metrics demonstrate strong performance for models finetuned on in-domain data, in OTS pretrained models they may not be well-calibrated to severely out-of-distribution (OOD) data like the noisy BPC domain.
% uses STAR scores, derived from confidence and ,  to re-weight each pseudo-label token during cross-entropy loss.
%Both approaches leverage i Whisper and Qwen3-ASR perform very poorly off-the-shelf (OTS) on the BPC domain, warranting an assessment of internal confidence metrics as pseudo-label filters for severely out-of-distribution (OOD) BPC audio. 
We conduct our study using two pretrained ASR models (Whisper and Qwen3-ASR) to pseudo-label BPC corpora from two U.S. cities: Baltimore, Maryland and Chicago, Illinois. We evaluate internal model confidence metrics \cite{kahn2020self,hu2024selftaughtrecognizerunsupervisedadaptation} as filtering mechanisms and contrast against an external LLM-as-a-judge filter that we design. The latter approach evaluates the contextual plausibility of generated transcripts and discards hypotheses that seem unlikely to occur within a policing domain.

To further explore the opportunities and limits of pseudo-labeling for BPC, we also assess two alternate pseudo-labeling paradigms: iterative pseudo-labeling (IPL) and cross-model pseudo-labeling. 
IPL has been successful on other domains by repeatedly pseudo-labeling the same set of training data with an improving model \cite{xu2020iterative,Likhomanenko2020slimIPLLI}. To our knowledge, this method has not been applied to extremely noisy settings where starting pseudo-labels have very high WER. We also introduce a new cross-model pseudo-labeling framework which uses pseudo-labels to enable cross-model training between Whisper and Qwen3-ASR.

% \af{We need to more generously acknowledge prior work on pseudolabeling here and make clear how our work is different. Here's a rough idea: Psuedolabeling can effectively improve ASR systems in domains with relatively clean data, where the pseudo-labels are high-quality (cite). These results have largely been conduct with X kinds of models, and the viability of pseudolabels with newer models and noisier data has not been investigated.}

% \af{In contrast, we address this gap by constructing pseudo-label pipelines focused on BPC, using strong-performing ASR systems as starting points (Whisper, Qwen3-ASR), and comparing approaches for filtering low-quality psuedo-labels before fine-tuning}

Our primary contributions are as follows:
\begin{itemize}
\item We show that pseudo-labeling unlabeled BPC audio can significantly improve pretrained ASR systems' performance on the BPC domain 
\item We demonstrate superior performance of   LLM-as-a-judge filtering for BPC pseudo-labels relative to internal confidence filtering metrics 
\item We identify cross-model pseudo-labeling as a promising future direction for pseudo-labeling as a domain adaptation strategy
\end{itemize}

\section{Related Work}

\subsection{ASR for Police Audio}

Calls to improve transparency and accountability in policing have motivated investigations of police data. Most research has focused on applying speech processing to footage from body-worn cameras \cite{voigt2017language,camp2021thin,field2023developing,rosas2025constructing}, with one line of work focusing on police radio communications from just the city of Chicago \cite{srivastava2024speech,narayanan2024race}.
Both body-worn camera and police radio footage suffer from poor audio quality and vocabulary specific to the policing domain, leading to poor performance of off-the-shelf ASR models  \cite{field2023developing,srivastava2024speech,rosas2025constructing}. While finetuning with manual transcriptions can improve performance \cite{field2023developing,srivastava2024speech,rosas2025constructing}, collecting manual transcriptions for this noisy, domain-specific  data is time-consuming and expensive. Furthermore, our preliminary experiments in  Table \ref{tab:cross-city-ft} in Appendix E demonstrated that training ASR systems on data from a different city can lead to degradation rather than improvements, suggesting manual transcriptions would need to be collected in every city of interest.
These challenges motivate our focus on pseudo-labeling to improve off-the-shelf performance. To the best of our knowledge, no work has leveraged unlabeled data to improve ASR on police audio. 
  
% While BWC and BPC audio contain different challenging characteristics (BWC contains overlapped speech and speakers are often far from the microphone while BPC is characterized by very short utterances), manual transcripts are difficult to obtain for both forms. It is reasonable then that any unsupervised domain adaptation method which improves BPC transcription would also benefit BWC transcription. 

\subsection{Pseudo-Labeling}
% \note{not sure how to write this section without it repeating a significant part of the intro}
Early ASR pseudo-labeling methods typically involved semi-supervised learning. In these approaches, a teacher model trained on a small amount of human-labeled audio generated transcripts for a larger unlabeled corpus, after which a student model was trained on the union of gold and pseudo-labeled data \cite{kahn2020self, manohar2021kaizen, higuchi2021momentum, 10832173}. Variants of this paradigm include continued training of the original teacher model on its own pseudo-labels rather than training a separate student \cite{chen2020semi}, as well as iterative pseudo-labeling methods that repeat pseudo-label generation and fine-tuning across multiple rounds \cite{xu2020iterative,Likhomanenko2020slimIPLLI}. Unlike our work, these previous approaches begin with an initial amount of labeled data to facilitate training of the teacher model. Log-probabilities from the teacher model initialized with supervised in-domain data served as effective quality indicators to filter pseudo-labels \cite{kahn2020self}.

The advent of general-purpose, pretrained ASR models offer opportunities to use pseudo-labeling as a fully unsupervised domain adaptation technique: a pretrained model can generate pseudo-labels without the need for in-domain gold data for initialization. \cite{hu2024selftaughtrecognizerunsupervisedadaptation} conduct an initial exploration using Whisper, and they further introduce weights derived from attention scores to account for potentially low-quality pseudo-labels. 
However, for the majority of datasets that \cite{hu2024selftaughtrecognizerunsupervisedadaptation} applies pseudo-labeling, the base Whisper model can already produce strong transcripts with WERs below 20; only on RATS \cite{walker2012rats}, and LS-FreeSound babble \cite{prasad2021investigation} does the base model perform significantly worse.  Thus, while internal metrics like log-probabilities and attention scores can distinguish between low and high-quality pseudo-labels on in-distribution data \cite{kahn2020self, hu2024selftaughtrecognizerunsupervisedadaptation}, the generalizability of these metrics to out-of-distribution (OOD) data like BPC remains unknown. 

\cite{10832173} explored external pseudo-label assessment by using an LLM-as-a-judge to identify high-quality monolingual pseudo-labels to improve a code-switched ASR system. This study demonstrated initial viability of LLM-based pseudo-label filtering, but it also relied on supervised training for model initialization, and the data used was acoustically clean \cite{Shi2020TheA2, panayotov2015librispeech,du2018aishell2transformingmandarinasr}. In pursuit of unsupervised adaptation to noisy domains, we apply LLM pseudo-label judging to BPC pseudo-labels from pretrained ASR systems. We identify contextual plausibility as a quality heuristic and leverage an external LLM to assess how much sense a given pseudo-label makes in the context of city-specific BPC communication. We compare the efficacy of this filtering approach to aforementioned internal metrics.

\section{Data and Models}
We apply our experiments to two BPC corpora from the cities of  Baltimore and Chicago \cite{morgan2025police, srivastava2024speech}. All work was conducted under protocols HIRB00019267 and HIRB00022007 approved by the Johns Hopkins Homewood Institutional Review Board. As reliable data de-identification is not possible for data of this volume,  we store the data on secure disk-encrypted servers and restrict access to IRB-approved team members.   

The Baltimore BPC dataset was recorded from 2016 to 2022 and covers the nine districts of the Baltimore Police Department. 
Audio from Friday evenings in 2017 were selected for transcription with rationale that Friday evenings contain higher amounts of police activity, and university students transcribed the audio \cite{fnlo}. 
We create our train and test splits by holding out samples from two days -- 2017-10-27 and 2017-06-02 -- for our test set of 2.29 hours. We create our train and validation sets from the remaining data with durations of 23.76 hours and 1.26 hours respectively.
This data has not previously been used for speech processing research. 

The Chicago BPC dataset consists of 12-months of audio, compiled using publicly available recordings from Broadcastify (\url{https://broadcastify.com}), covering 11 of the 13 Chicago Police Department dispatch zones. 
% In total, 46.2 hours of speech were transcribed, of which 42.5\% were labeled by researchers at the University of Chicago, and the remaining 57.5\% were labeled through a professional transcription service. The initial train, validation, and test splits were obtained by stratified sampling in 22 strata comprising 11 dispatch zones and two time periods (day/night). To evaluate model robustness in unfamiliar contexts, utterances from zones 5 and 6 were excluded from training and split evenly between the validation and test sets. The remaining nine zones followed the standard 80\% train, 10\% validation, and 10\% test split, aggregating to 33.0 hours, 6.6 hours, and 6.6 hours respectively \cite{srivastava2024speech}.
To match durations of our Baltimore BPC dataset, we randomly subsample the original train-validation-test splits from \cite{srivastava2024speech} to obtain a training set of 23.89 hours, a validation set of 1.25 hours, and a test set of 2.25 hours.

We make use of Whisper large-v3 \cite{radford2022robustspeechrecognitionlargescale} and Qwen3-ASR3-ASR 1.7B \cite{shi2026qwen3} as our state-of-the-art, open-source, pretrained ASR models of choice. Hyperparameters used for finetuning are in Appendix \ref{app:model_training}.

% Our starting set of untranscribed data has a duration of ~83 hours. We remove samples where the audio is less than a second, and we produce pseudolabels by running inference on this set with both the Whisper and Qwen3-ASR base models. After inference, we discard samples where the pseudolabel contains eight or more consecutive repetitions of an n-gram to remove obvious instances where the model got stuck in a loop during decoding. We also discard samples where the pseudolabel contains no text.   

% We obtain a 24 hour subset of pseudolabeled data by randomly selecting audio samples to run inference on from the larger set of untranscribed data. We apply the same empty label and n-gram filtering to this dataset.  

% We create our mixed dataset by randomly selecting five hours worth of samples from the transcribed dataset and shuffling these with the 24 hour subset of pseudolabels.

\section{Methods}
\subsection{Pseudo-labeling pipeline}
We construct pipelines for fully unsupervised pseudo-labeling. Starting with an off-the-shelf (OTS) ASR model, we generate pseudo-labels for audio files, filter the pseudo-labels, and  then fine-tune the same model using the retained set of pseudo-labeled data. We empirically investigate five methods for filtering out low-quality pseudo-labels before fine-tuning:  two approaches (unfiltered and oracle) serve as references, two are derived from prior work (log probabilities and STAR scores), and one (LLM-as-a-judge) we propose in this work.

\textbf{Unfiltered} We acquire pseudo-labels by running inference on the training split audio from the Baltimore and Chicago BPC datasets. We discard samples where the pseudo-label contains n-grams between five and eight words long repeating more than eight times to filter out obvious instances of looping. 

\textbf{Oracle}
We use human-written transcripts to calculate WER for each set of pseudo-labels and discard samples with WER greater than 0.2. WERs presumably would not be known in a real setting without human transcripts, but we conduct oracle-filtering to examine if pseudo-label finetuning can improve ASR performance under perfect filtering conditions and to establish a ceiling on how much improvement we can expect to achieve with pseudo-labels.

\textbf{Log-probability}  We calculate the log-probability of each pseudo-label by averaging the log-probabilities of the OTS ASR model for each token in the pseudo-label. We filter out all samples where the pseudo-label has an average log-probability less than -1. We base this threshold off of Whisper's decoding strategy which increases temperature and re-runs decoding when the average log-probability of the generated tokens is lower than -1 \cite{radford2022robustspeechrecognitionlargescale}. 

% \cite{kahn2020self} finds filtering critical for successful pseudo-label finetuning, improving ASR for noisy LibriSpeech by using log-probabilities to discard the bottom 30\% of pseudo-labels. This study generated pseudo-labels from a base model trained from scratch on labeled in-domain data. We examine if log-probability filtering still works when pretrained models generate pseudo-labels for out-of-domain, very noisy data. 

\textbf{STAR score} We apply the Self-Taught Recognizer (STAR) method from \cite{hu2024selftaughtrecognizerunsupervisedadaptation} to calculate token-level probabilities and attentive scores from OTS models and use these to  reweight the cross-entropy loss calculation during finetuning. Samples are fully discarded from the training set when the transcribing model produces no output tokens. All other samples are retained, and individual tokens in the sample receive higher or lower STAR scores depending on the attentive-score and posterior probability of the token.

% The Self-Taught Recognizer (STAR) approach uses confidence and attention scores  to re-weight each pseudo-label token according to its estimated correctness during cross-entropy loss \cite{hu2024selftaughtrecognizerunsupervisedadaptation}. The datasets in which STAR enables successful pseudo-label finetuning all have OTS WERs below 20, with only CORAAL \cite{kendall2023coraal}, RATS \cite{walker2012rats}, and LS-FreeSound babble \cite{prasad2021investigation} having WERs of 21.5, 46.9 and 40.2 respectively. Qwen3-ASR and Whisper OTS WER on BPC is substantially higher than these numbers, motivating an investigation of whether token-level reweighting can overcome very poor initial pseudo-labels. 

\textbf{LLM-as-a-judge}
% Previously mentioned pseudo-label filtering methods like log-probabilities or STAR scores \cite{hu2024selftaughtrecognizerunsupervisedadaptation} are computed autoregressively, token-by-token by the generating model. They do not take a global perspective on the transcription's overall coherence. Global coherence is a potential heuristic to identify incorrect pseudo-labels, as some transcripts obviously do not make sense in the context of policing. We address this oversight by using an an external LLM-as-a-judge \cite{zheng2023judging}  to filter BPC pseudo-labels. With extensive world knowledge and contextual reasoning capabilities, LLMs can evaluate the semantic plausibility of a pseudo-label given the context of city-specific BPC audio. 
We prompt an LLM with the situational and locational context of the audio (eg. police radio from Baltimore)  and each generated pseudo-label, instructing the model to judge if the pseudo-label seems accurate given the context. The full prompt is in Appendix \ref{app:standard_prompt}. Samples which the LLM marks as inaccurate are dropped. For our primary experiments, we use Meta-Llama-3-8B-Instruct \cite{meta2024introducing} as the judge LLM. Unlike log-probabilities and STAR scores, this approach leverages external  knowledge (model parametric knowledge) to score pseudo-label plausibility.

\subsection{Pipeline ablations}
We conduct two ablations to evaluate our LLM-as-a-judge filter. In the first, we replace Meta-Llama-3-8B-Instruct with Meta-Llama-3.3-70B-Instruct to assess the effect of a larger judge model. In the second, we construct a finer-grained, more informative prompt which includes the duration of the audio sample and instructions to judge whether the length of the proposed transcript is consistent with the length of the audio. The prompt also includes explicit rules for judging, obtained by prompting Gemini-3.5-Flash \cite{team2023gemini} to analyze errors where the baseline LLM misjudged high-WER pseudo-labels as correct. We share this prompt in Appendix \ref{app:rule-based_prompt}. These rules are derived from a particular set of pseudo-labels, and we evaluate the new prompt on a newly generated set of pseudo-labels. Although this prompt relies impractically on knowing true WERs, we use it to estimate an upper bound on gains achievable from prompt engineering.

% We conduct ablation experiments to explore robustness and potential limits of our methods. For our LLM-as-a-judge filtering method, we conduct two ablations. First, we exchange Meta-Llama-3-8B-Instruct for Meta-Llama-3-70B to examine the effect of a larger judge model. In the second, we construct a finer-grained, more informative prompt for the judge model. The refined prompt includes information about the audio, specifically the audio duration in seconds as well as instructions to judge whether the length of the proposed transcript is consistent with the length of the audio. We also modify the LLM prompt to provide explicit rules for judging. We develop these rules by passing a file of pseudo-labels with LLM-judgements and their WERs to Gemini-3.5-Flash \cite{team2023gemini}. We prompt the model to find patterns in the samples where the LLM judged the pseudo-label as correct even though the WER was high, and to propose a rule-based LLM prompt for better filtering. We share our rule-based prompt in Appendix \ref{app:rule-based_prompt}. Since these rules include examples specific to the set of pseudo-labels we provided to the Gemini model, we generate a new set of pseudo-labels to use with this prompt in our ablation. While generating rules in this way is not directly reproducible in practical settings where true WER is not known, this experiment offers a strong indicator of how much benefit may be achievable through further prompt engineering.

We also conduct oracle filtering ablations with higher WER thresholds to examine if consistent but less stringent pseudo-label filters can still enable substantial performance gains.

\subsection{Alternate Pseudo-Labeling Paradigms}
Finally, we investigate alternate unsupervised  ways that pseudo-labels might adapt pretrained ASR models to the BPC domain.

Theoretically, pseudo-labeling may not improve a model because it does not present new speech-to-audio mappings. To circumvent this, we fine-tune Qwen3-ASR and Whisper with LLM-judged pseudo-labels from the other model. We call this `cross model pseudo-label finetuning'. We also implement IPL  (iterative pseudo-labeling), where after one round of standard pseudo-label finetuning, we use  the model to generate another set of pseudo-labels which are used for even further finetuning.  We conduct three rounds of IPL with LLM-as-a-judge filtering, re-labeling and re-judging the entire training split after each fine-tuning step.

\subsection{Normalization}
 WER can differ significantly depending on how the ground truth and hypothesis text are normalized \cite{kuhn2024measuring, radford2022robustspeechrecognitionlargescale}. Our normalization scheme applies the Whisper English text normalizer, then targets the spoken number conventions common in police radio communication. Spoken number words are converted to their digit equivalents, compound numbers are resolved with word-to-number conversion, and spoken pairs are mapped to their digit form. We then strip punctuation, space-separate digits, and normalize whitespace. A key motivation for this normalization pipeline is the prevalence of 10-codes, standard numeric radio codes (e.g., ``10-4'', ``10-96'') that ASR systems can transcribe into various forms (e.g., ``ten four'', ``10-4'', or ``104''). We apply the same normalization to both reference transcriptions and system hypotheses prior to WER computation. Because this normalization substantially reduces WER, we report the un-normalized scores for the experiments from Table \ref{tab:horizontal_main_results} in Tables \ref{tab:Qwen3-ASR_unnorm}, and \ref{tab:whisper_unnorm} in Appendix D. 
 
% \subsection{Significance Testing}
% We operate the models deterministically with temperature = 0. For filtering experiments, we report the mean of three finetuning runs, evaluating statistical significance ($\alpha=0.05$) against the OTS baseline via a one-sample t-test and against the unfiltered baseline via Welch’s t-test.

% As each model is operated with a temperature of 0, our OTS results are deterministic. We conduct three runs of each filtering  experiment, and we use a one-sample t-test to calculate statistical significance between each of the five filtering methods and the OTS baseline. We use Welch's t-test to calculate significance between oracle-filtering, log-probability filtering, STAR scores, and LLM-as-a-judge with the unfiltered result as a baseline. We use $\alpha = 0.05$ in both significance tests.  

\section{Results}

\begin{table*}[htbp]
\caption{Different pseudo-label filtering approaches applied to Chicago and Baltimore BPC corpora on the Qwen3-ASR and Whisper ASR models. * indicates statistical significance relative to the OTS baseline, and $^{\dagger}$ indicates statistical significance relative to the unfiltered result.}
\label{tab:horizontal_main_results}
\centering
\fontsize{8pt}{10pt}\selectfont      % Crisper, perfectly proportional text scale
\setlength{\tabcolsep}{3.8pt}       % Controlled horizontal padding to lock table inside margins
\begin{tabular}{lcccccccccccc}
\toprule
& \multicolumn{6}{c}{\textbf{Baltimore}} & \multicolumn{6}{c}{\textbf{Chicago}} \\
\cmidrule(lr){2-7} \cmidrule(lr){8-13}
& \multicolumn{3}{c}{\textbf{Qwen3-ASR}} & \multicolumn{3}{c}{\textbf{Whisper}} & \multicolumn{3}{c}{\textbf{Qwen3-ASR}} & \multicolumn{3}{c}{\textbf{Whisper}} \\
\cmidrule(lr){2-4} \cmidrule(lr){5-7} \cmidrule(lr){8-10} \cmidrule(lr){11-13}
\textbf{Method} & \textbf{\makecell{Train\\Hours}} & \textbf{\makecell{Train\\WER}} & \textbf{\makecell{Test\\WER}} & \textbf{\makecell{Train\\Hours}} & \textbf{\makecell{Train\\WER}} & \textbf{\makecell{Test\\WER}} & \textbf{\makecell{Train\\Hours}} & \textbf{\makecell{Train\\WER}} & \textbf{\makecell{Test\\WER}} & \textbf{\makecell{Train\\Hours}} & \textbf{\makecell{Train\\WER}} & \textbf{\makecell{Test\\WER}} \\ \hline 
OTS              & -      & -        & 0.3611            & -      & -        & 0.3690            & -      & -        & 0.3439            & -      & -        & 0.4351            \\ \hline
Oracle-filter    & 5.56    & 0.0987   & 0.3276*$^{\dagger}$ & 6.25  & 0.0947   & 0.3517*$^{\dagger}$ & 9.22   & 0.1142   & 0.2699*$^{\dagger}$ & 7.31   & 0.1445   & 0.3567*$^{\dagger}$ \\ \hline
Unfiltered       & 23.70  & 0.3943   & 0.3597            & 23.76  & 0.4113   & 0.3856            & 23.89  & 0.3556   & 0.3153* & 23.89  & 0.6176   & 0.3986            \\ 
Log-prob         & 23.64  & 0.3936   & 0.3562* & 23.14  & 0.4027   & 0.3757* & 23.88  & 0.3593   & 0.3146* & 22.54  & 0.4130   & 0.3935* \\ 
STAR             & 23.63  & 0.3889   & 0.3719            & 23.76  & 0.4037   & 0.4317            & 22.31  & 0.3341   & 0.3194*            & 24.00  & 0.4251   & 0.4319            \\ 
LLM-as-judge     & 16.09  & \textbf{0.3340} & \textbf{0.3496}*$^{\dagger}$ & 15.83  & \textbf{0.3401} & \textbf{0.3685}$^{\dagger}$ & 18.17  & \textbf{0.2880} & \textbf{0.3051}*$^{\dagger}$ & 16.31  & \textbf{0.3154} & \textbf{0.3905}    \\ \hline
\end{tabular}

\vspace{2pt}
\begin{minipage}{0.95\linewidth}
\footnotesize
\textit{Note:} We operate the models deterministically with temperature = 0. For filtering experiments, we report the mean of three finetuning runs, evaluating statistical significance ($\alpha=0.05$) against the OTS baseline via a one-sample t-test and against the unfiltered baseline via Welch’s t-test.
\end{minipage}

\end{table*}

Table \ref{tab:horizontal_main_results} reports pseudo-labeling performance under all five filtering methods as well as OTS baselines. Pseudo-label finetuning achieves significant improvements for both models on both datasets, but results vary depending on the filtering method applied. LLM-as-a-judge filtering consistently provides the greatest improvements, reducing WER of Qwen3-ASR, the stronger of the two models, from 0.3439 on the Chicago corpus to 0.3051 (11.28\% WER reduction) and from 0.3611 on the Baltimore corpus to 0.3496 (3.18\% WER reduction). 

Log-probability filtering removes very few samples, discarding less than two hours of data in each experiment. This consistently creates a train set with marginally lower WER than the unfiltered train set, except Whisper pseudo-labels on the Chicago dataset where the difference is substantial. Log-probability filtering inconsistently improves on the OTS test set performance, but these improvements are not significant relative to the unfiltered pseudo-labels. The witnessed improvements may be a product of the pseudo-labels themselves rather than of the filtering method. 

STAR discards more data, and the retained pseudo-label train set has lower WER than the log-probability filtered train set. However, token-level STAR score reweighting does not significantly improve test set WER in relation to the OTS model performance. In three out of four cases, test set WER from the STAR experiment is \textit{higher} than the OTS result. 

Relative to the previous two methods, LLM-as-a-judge filtering consistently discards more data and obtains the lowest train set WER. With the exception of Whisper on the Chicago test set, this filtering does translate to statistically significant improvements relative to the OTS result and the unfiltered pseudo-labels. This exceptional case may be tied to Whisper's high starting WER on the Chicago test set. While LLM-judging filters best out of the three examined filtering methods, results still lag relative to an oracle-filter

\subsection{Oracle Filtering}
Oracle filtering with WER threshold of 0.2 achieves the lowest test set WER across all methods. Compared to OTS baselines, it reduces Baltimore test WER by 4.7\% for Whisper and 9.3\% for Qwen3-ASR, and Chicago test WER by 20.3\% and 22.0\%, respectively. Oracle filtering discards roughly twice as much data as the LLM-as-a-judge filter and reduces test set WER by as much as 11.5\% relative to the LLM-judged result.
% Oracle filtering achieves significantly lower test set WER than all other examined filtering methods. On the Baltimore test set, a WER threshold of 0.2 attains a 4.7\% WER decrease and 9.3\% WER decrease for Whisper and Qwen3-ASR relative to each model's OTS performance. The same threshold enables even greater improvements on the Chicago data with WER dropping by 20.3\% for Whisper and by 22.0\% for Qwen3-ASR. Oracle filtering discards roughly twice as much data as LLM-as-a-judge filtering, and test set performance can be as much as 11.5\% lower than LLM-judged test set performance. 
Discarding samples with WER over 0.2 without access to human transcripts is challenging and potentially unrealistic for such noisy domain data. Table \ref{tab:oracle_thresholds} shows that looser WER thresholds of 0.3 and 0.4 still enable improvements in Qwen3-ASR relative to the OTS baseline across both corpora. 

\begin{table}[htbp]
\caption{Qwen3-ASR Finetuned with Oracle-Filtered Pseudo-labels with different WER thresholds.}
\label{tab:oracle_thresholds}
\centering
\fontsize{8.5pt}{10pt}\selectfont
\begin{tabular*}{\columnwidth}{l@{\extracolsep{\fill}}cccccc}
\toprule
& \multicolumn{3}{c}{\textbf{Baltimore}} & \multicolumn{3}{c}{\textbf{Chicago}} \\
\cmidrule(lr){2-4} \cmidrule(lr){5-7}
\multicolumn{1}{l}{\textbf{\makecell{WER\\Thrsh.}}} & \textbf{\makecell{Train\\WER}} & \textbf{\makecell{Train\\Hrs}} & \textbf{\makecell{Test\\WER}} & \textbf{\makecell{Train\\WER}} & \textbf{\makecell{Train\\Hrs}} & \textbf{\makecell{Test\\WER}} \\ \hline 
0.2 & 0.0987 & 5.56  & 0.3276 & 0.1142 & 9.22  & 0.2684 \\ \hline
0.3 & 0.1437 & 8.65  & 0.3314 & 0.1539 & 12.48 & 0.2791 \\ \hline
0.4 & 0.1836 & 11.45 & 0.3387 & 0.1858 & 14.94 & 0.2860 \\ \hline
\end{tabular*}
\end{table}

\subsection{LLM-as-a-judge Ablations}

\begin{table}[htbp]
\caption{LLM-as-a-judge Ablations: Judge Model Variation and Rule-Based Prompting Applied to Qwen3-ASR}
\label{tab:llm-judge-ablations}
\centering
%\begin{tabular}{0.9\columnwidth}{cccc}
\begin{tabular}{cccc}
\toprule
\textbf{\makecell{Judging \\Model}} & \textbf{\makecell{Prompt\\Type}} & \textbf{\makecell{Baltimore \\Test WER}} & \textbf{\makecell{Chicago\\ Test WER}} \\
\midrule
Llama 8B & Standard & 0.3496 & 0.3051
\\ \hline
Llama 8B & Rule-based & 0.3436 & 0.3000 \\
Llama 70B & Standard & 0.3575 & 0.3050 \\
\bottomrule
\end{tabular}
\end{table}

\begin{table}[htbp]
\caption{Cross-Model Pseudo-Label Finetuning}
\label{tab:cross_model_finetuning}
\centering
\begin{tabularx}{\columnwidth}{
  >{\raggedright\arraybackslash\hsize=0.95\hsize}X 
  >{\centering\arraybackslash\hsize=0.95\hsize}X 
  >{\centering\arraybackslash\hsize=1.05\hsize}X 
  >{\centering\arraybackslash\hsize=1.05\hsize}X
}
\toprule
\textbf{\makecell{Base\\Model}} & \textbf{\makecell{PL\\Model}} & \textbf{\makecell{Baltimore\\Test WER}} & \textbf{\makecell{Chicago\\Test WER}} \\
\midrule
\multirow{2}{*}{Qwen3-ASR}  & Qwen3-ASR & 0.3496 & 0.3051 \\
                            & Whisper   & 0.3365 & 0.3260 \\
\midrule
\multirow{2}{*}{Whisper}    & Whisper   & 0.3685 & 0.3905 \\
                            & Qwen3-ASR & 0.3546 & 0.3537 \\
\bottomrule
\end{tabularx}
\end{table}

% \begin{table}[htbp]
% \caption{Cross-Model Pseudo-Label Finetuning}
% \label{tab:cross_model_finetuning}
% \centering
% \begin{tabularx}{\columnwidth}{
%   >{\raggedright\arraybackslash\hsize=0.7\hsize}X 
%   >{\centering\arraybackslash\hsize=0.6\hsize}X 
%   >{\centering\arraybackslash\hsize=1.3\hsize}X 
%   >{\centering\arraybackslash\hsize=1.4\hsize}X
% }
% \toprule
% \textbf{\makecell{Base\\Model}} & \textbf{\makecell{PL\\Model}} & \textbf{\makecell{Baltimore Test\\Set WER}} & \textbf{\makecell{Chicago Test\\Set WER}} \\
% \midrule
% \multirow{\mbox{Qwen3-ASR}}    & Qwen3-ASR    & 0.3496 & 0.3051 \\ \cmidrule(lr){2-4}
%                          & Whisper & 0.3365 & 0.3260 \\ 
% \midrule
% \multirow{\mbox{Whisper}} & Whisper & 0.3685 & 0.3905 \\ \cmidrule(lr){2-4}
%                          & Qwen3-ASR    & 0.3546 & 0.3537 \\
% \bottomrule
% \end{tabularx}
% \end{table}

Table \ref{tab:llm-judge-ablations} shows that neither a larger judging model nor a rule-based prompt tailored to the domain majorly shrink the gap between LLM-judging and oracle filtering. Relative to the standard prompt, the rule-based prompt enables a 1.7\% WER improvement across both corpora. Using the larger Llama 70B model with our standard prompt worsens WER on the Baltimore test set by 2.3\% and improves performance on the Chicago test set by .03\%. These marginal performance variations suggest prompt engineering or choice of LLM are not promising directions for further improvements.

\subsection{Alternate Pseudo-labeling Paradigms}

\begin{table*}[t]
\caption{IPL with Qwen3-ASR on Baltimore and Chicago BPC}
\label{tab:IPL}
\centering
\begin{tabular}{llccccccccc}
\toprule
 & & \multicolumn{3}{c}{\textbf{Iteration 1}} & \multicolumn{3}{c}{\textbf{Iteration 2}} & \multicolumn{3}{c}{\textbf{Iteration 3}} \\
\cmidrule(lr){3-5} \cmidrule(lr){6-8} \cmidrule(lr){9-11}
\textbf{Model} & \textbf{City} & \textbf{\makecell{Train \\WER}} & \textbf{\makecell{Train \\Duration}} & \textbf{\makecell{Test \\WER}} & \textbf{\makecell{Train \\ WER}} & \textbf{\makecell{Train \\Duration}} & \textbf{\makecell{Test \\WER}} & 
\textbf{\makecell{Train \\ WER}} & \textbf{\makecell{Train \\Duration}} & \textbf{\makecell{Test \\WER}}  \\

\midrule
Qwen3-ASR & Baltimore  & 0.3340  & 16.09 hrs  & 0.3517  & 0.3552  & 18.06 hrs  & 0.3464  & 0.3660 & 18.89 hrs  & 0.3473  \\
Qwen3-ASR & Chicago  & 0.2886 & 18.15 hrs & 0.3042  & 0.3095 & 19.67 hrs & 0.3018 & 0.3307 & 20.59 hrs & 0.3013  \\
\bottomrule
\end{tabular}
\end{table*}

% \begin{table}[htbp]
% \caption{Cross-Model Pseudo-Label Finetuning}
% \label{tab:cross_model_finetuning}
% \centering
% \begin{tabularx}{\columnwidth}{
%   >{\raggedright\arraybackslash\hsize=0.7\hsize}X 
%   >{\centering\arraybackslash\hsize=0.6\hsize}X 
%   >{\centering\arraybackslash\hsize=1.3\hsize}X 
%   >{\centering\arraybackslash\hsize=1.4\hsize}X
% }
% \toprule
% \textbf{Base Model} & \textbf{PL Model} & \textbf{\makecell{Baltimore Test\\Set WER}} & \textbf{\makecell{Chicago Test\\Set WER}} \\
% \midrule
% \multirow{\mbox{Qwen3-ASR}}    & Qwen3-ASR    & 0.3496 & 0.3051 \\ \cmidrule(lr){2-4}
%                          & Whisper & 0.3365 & 0.3260 \\ 
% \midrule
% \multirow{\mbox{Whisper}} & Whisper & 0.3685 & 0.3905 \\ \cmidrule(lr){2-4}
%                          & Qwen3-ASR    & 0.3546 & 0.3537 \\
% \bottomrule
% \end{tabularx}
% \end{table}

IPL with LLM-judging provides very minor improvements compared to a single round of pseudo-labeling: in Table \ref{tab:IPL}, WER reduces by 1.25\%  on the Baltimore dataset and by 0.95\% on the Chicago dataset  between the first and third IPL iterations. With each iteration, the LLM-as-a-judge filter retains more pseudo-labels, and the train set has a higher overall WER, potentially indicating that IPL with LLM-judging leads to transcripts that look more plausible for the domain without necessarily increasing accuracy.

Table \ref{tab:cross_model_finetuning} shows that, relative to OTS performance, cross-model pseudo-label finetuning improves Whisper transcriptions across both test sets and improves Qwen3-ASR performance on the Baltimore test set. The OTS Qwen3-ASR model is substantially stronger on the Chicago data than the OTS Whisper model, and in this case finetuning the model on Whisper's Chicago pseudo-labels provides a worse result than self-training, potentially showing that noise in Whisper's pseudo-labels detracted from the value of providing new mappings to the Qwen3-ASR model. On the Baltimore corpus, where Whisper and Qwen3-ASR perform comparably, cross-model finetuning improves both models more than self-training with their own pseudo-labels.

\section{Analysis of Pseudo-labeling Improvements}
\begin{table}[htbp]
\caption{Words from the Baltimore and Chicago corpora where transcriptions improve with pseudo-label finetuning}
\label{tab:Qwen3-ASR-baltimore-error-analysis}
\centering
\setlength{\tabcolsep}{2.0pt} 
\begin{tabularx}{\columnwidth}{
  >{\centering\arraybackslash}X
  >{\centering\arraybackslash}X
  >{\centering\arraybackslash}X |
  >{\centering\arraybackslash}X
  >{\centering\arraybackslash}X
  >{\centering\arraybackslash}X
} 
\toprule
% FIXED: Changed \multicolumn span from 2 to 3 to span the entire section
\multicolumn{3}{c}{\textbf{Baltimore}} & \multicolumn{3}{c}{\textbf{Chicago}} \\
\midrule
% \cmidrule(lr){1-3}\cmidrule(lr){4-6}
\textbf{Word} & \textbf{\makecell{$\Delta$ Error\\Rate}}& \textbf{\makecell{Word\\Count}} & \textbf{Word} & \textbf{\makecell{$\Delta$ Error\\Rate}} & \textbf{\makecell{Word\\Count}} \\ 
\midrule
the  & -1.84  & 543 & robert & -7.08 & 376 \\
charlie  & -5.88  & 170 & 104 & -4.77 & 419 \\
in  & -2.79 & 215 & on & -4.12 & 364 \\
copy  & -16.66 & 36 & we & -4.86 & 247 \\
i  & -1.30 & 461 & you & -2.31 & 520 \\
that & -2.61 & 191 & ahead & -23.81 & 42 \\
32  & -7.69 & 65 & go & -9.19 & 98 \\
go  & -10.00 & 50 & okay & -7.38 & 122 \\
going  & -4.42 & 113 & and & -2.13 & 423 \\
am & -3.44 & 145  & are & -4.38 & 183 \\
c & -8.69 & 46 & a & -1.26 & 557 \\
11 & -7.69 & 52 & call & -9.86 & 71 \\
with & -3.50 & 114  & squad & -8.98 & 78 \\
them & -11.42 & 35  & 31 & -30.00 & 20 \\
complainant & -22.22 & 18 & 4 & 21.43 & 28 \\
to & -0.83 & 479 & to & -1.73 & 347 \\
it & -2.04 & 196 & park & -24.00 & 25 \\
one & -2.41 & 124  & one & -6.00 & 100 \\
can & -2.11 & 142 & stop & -10.00 & 60 \\
reference & -5.66 & 53 & all & -5.31 & 113 \\
\bottomrule
\end{tabularx}
\end{table}
Finally, we analyze how finetuning with pseudo-labels improves Qwen3-ASR transcriptions on the BPC domain.  
Table \ref{tab:Qwen3-ASR-baltimore-error-analysis}
shows the 20 terms whose transcriptions improve most from pseudo-label fine-tuning. To attain this list, we normalize the OTS, finetuned, and ground truth transcripts with the Whisper normalizer. For each word in the ground-truth transcripts, we subtract its error rate in the off-the-shelf model transcripts ($P^{\text{incorrect}}_{\text{OTS}}$) from its error rate in the fine-tuned model transcripts ($P^{\text{incorrect}}_{\text{FT}}$). We scale this quantity by the word's relative frequency in the ground truth corpus to prioritize frequently occurring terms:
% these words by ranking words from the set of words from the reference transcripts according to how their transcripts improve with pseudo-label finetuning. We describe the ranking calculating in Equation \ref{eq:improvement_score}:
\begin{equation}
\label{eq:improvement_score}
R = (P^{\text{incorrect}}_{\text{FT}} - P^{\text{incorrect}}_{\text{OTS}}) \times freq
\end{equation}
We select the 20 words with the most negative $R$ values from each corpus for Table \ref{tab:Qwen3-ASR-baltimore-error-analysis}.

The finetuned model recognizes a number of policing-specific words better, including words from the policing phonetic-alphabet (eg. ``charlie'' and ``robert'') \cite{bmore-radio-comms}, and numbers used for police 10-codes (eg. 10-4, 10-31, 10-32, 10-11) \cite{departmental-radio-comms}, suggesting that the model does successfully adapt to domain-specific vocabulary.
Fine-tuning also improves performance over frequent function words (e.g., ``the'', ``that'', ``and''), implying that overall model quality improves as well.

% To show the practical impact of fine-tuning with LLM-judged pseudo-labels,
We also compare sample-level WERs between the off-the-shelf and fine-tuned Qwen3-ASR models by subtracting the fine-tuned WER from the baseline WER for each sample.  The top five samples per corpus with the most positive differences are presented in Table \ref{tab:combined_qualitative_samples}.
% highlight the greatest transcription improvements gained from finetuning with pseudo-labels.
There is an observable improvement in numeric transcriptions across both corpora, which are frequently used in police communication and transmit important information: 10-codes are used to communicate about vehicle stops, crimes in progress, body-worn camera usage, etc. Successfully transcribing these codes offers broad visibility into policing activity that is not observable in other department-reported statistics \cite{morgan2025police, ColePPI}. Overall, the examples in Table \ref{tab:combined_qualitative_samples} highlight that finetuning with LLM-judged pseudo-labels improves performance for domain-relevant content.
 % We hypothesize that when given the policing context, the LLM judge favors numbers as transcripts over close homophones (ie. eight vs. wait), and this pushes the model to transcribe audio as numbers.

\begin{table}[htbp]
\caption{Samples from Chicago and Baltimore BPC corpora where pseudo-label finetuning improves on OTS transcripts}
\label{tab:combined_qualitative_samples}
\centering
\fontsize{8.5pt}{10.5pt}\selectfont
\begin{tabular*}{\columnwidth}{l@{\extracolsep{\fill}}ll}
\toprule
\multicolumn{3}{c}{\small\textbf{Chicago}} \\
\cmidrule(lr){1-3}
\fontsize{7.5pt}{9pt}\selectfont\textbf{\makecell[l]{Ground\\Truth}} & 
\fontsize{7.5pt}{9pt}\selectfont\textbf{\makecell[l]{OTS\\Prediction}} & 
\fontsize{7.5pt}{9pt}\selectfont\textbf{\makecell[l]{FT\\Prediction}} \\ \midrule
2471 & it is your 1st time i & 2475 \\ \hline
935 & nice to meet you & 935 \\ \hline
771 & i do not have anyone & 71 \\ \hline
104 & i am looking for & 104 \\ \hline
104 & what you are paying for & 104 \\ \midrule 
\multicolumn{3}{c}{\small\textbf{Baltimore}} \\
\cmidrule(lr){1-3}
\fontsize{7.5pt}{9pt}\selectfont\textbf{\makecell[l]{Ground\\Truth}} & 
\fontsize{7.5pt}{9pt}\selectfont\textbf{\makecell[l]{OTS\\Prediction}} & 
\fontsize{7.5pt}{9pt}\selectfont\textbf{\makecell[l]{FT\\Prediction}}  \\ \midrule
citywide & \textchinese{阻我}citywide & citywide \\ \hline
612nd & 6 more seconds & 612nd \\ \hline
yessir adam & \makecell[l]{i do not know what\\else we got} & i do not know \\ \hline
92 & \makecell[l]{magnitude you are\\showing us you are\\showing here} & \makecell[l]{magnitude showing a\\seastorm here} \\ \hline
charlie 22 & are they playing too  & charlie 22 \\ 
\bottomrule
\end{tabular*}
\end{table}

% \begin{table}[htbp]
% \caption{Chicago BPC Samples where the Qwen3-ASR model finetuned on LLM-judged pseudo-labels transcribed better than the off-the-shelf model}
% \label{tab:chicago_examples}
% \centering
% \begin{tabularx}{\columnwidth}{XXX}
% \toprule
% \textbf{Ground Truth} & \textbf{OTS Prediction} & \textbf{\makecell{Finetuned Model\\Prediction}} \\
% \midrule
% 2471 & it is your 1st time i & 2475 \\ \hline
% 935 & nice to meet you & 935 \\ \hline
% 771 & i do not have anyone & 71 \\ \hline
% 104 & i am looking for & 104 \\ \hline
% 104 & what you are paying for & 104  \\
% \bottomrule
% \end{tabularx}
% \end{table}

% \begin{table}[htbp]
% \caption{Baltimore BPC Samples where the Qwen3-ASR model finetuned on LLM-judged pseudo-labels transcribed better than the off-the-shelf model}
% \label{tab:baltimore_examples}
% \centering
% \begin{tabularx}{\columnwidth}{XXX}
% \toprule
% \textbf{Ground Truth} & \textbf{OTS Prediction} & \textbf{\makecell{Finetuned Model\\Prediction}} \\
% \midrule

% citywide & \textchinese{所以 我} & citywide \\ \hline
% 612nd & 6 more seconds & 612nd \\  \hline
% yessir adam & i do not know what else we got & i do not know \\ \hline
% 92 & magnitude you are showing us you are showing here  & magnitude showing a seastorm here \\  \hline
% charlie 22 & are they playing too & charlie 22 \\
% \bottomrule
% \end{tabularx}
% \end{table}

\section{Discussion}
Our filtering experiments reveal an interesting challenge to pseudo-labeling noisy, OOD data with foundation ASR models. Internal confidence metrics that effectively filter in-domain pseudo-labels produce middling results on BPC while external knowledge-based filtering consistently succeeds. This suggests that despite their extensive pretraining, these foundation models' internal confidence metrics may not be calibrated to noisy, specialized domains. 

The substantial improvement that LLM-judged pseudo-label finetuning achieves over BPC-specific terms is also meaningful despite final BPC WERs being higher than on standard benchmarks like LibriSpeech \cite{panayotov2015librispeech}. These improvements over numeric terms and police alphabet make analysis of BPC main topics more feasible.

Our oracle filtering experiment also shows that even greater gains are possible with  better filtering. 
% The LLM-as-a-judge filter exploits domain-specific lexical patterns that distinguish police communication from general speech. 
Modest improvements from our ablations suggest that the LLM-as-a-judge filter could already be close to its practical ceiling when judging contextual plausibility of the transcript text alone. 
% This is potentially a broader strategy for specialized domain adaptation. 
The remaining gap to oracle filtering may reflect errors beyond contextual plausibility, making a case for combined filtering approaches that also account for uncertainty due to noise in the audio signal. 

Cross-model pseudo-labeling suggests an important relationship between model quality and model diversity that should be explored in future pseudo-labeling work. While self-training with LLM-judged pseudo-labels aims to promote a model's strengths while minimally reinforcing poor transcriptions, cross-model finetuning may have the added benefit of \textit{correcting} a model's weaknesses. Qwen3-ASR and Whisper both benefit from cross-model training on Baltimore BPC where both have similar OTS performance, but the same paradigm degrades Qwen3-ASR when trained on Whisper's weaker pseudo-labels of the Chicago BPC corpus. Future work should investigate how similarly models must perform to reap the benefits of pseudo-labeling and whether a group of marginally weaker models can strengthen a main base model.

Beyond pseudo-labeling, unlabeled data can also be used to continue pretraining for self-supervised models \cite{chen_wavlm_2022, hsu_hubert_2021, baevski_wav2vec_2020}. While the base models commonly used for self-supervised training typically exhibit weaker OTS performance than Whisper or Qwen3-ASR, their ability to learn directly from unlabeled audio may prove advantageous to the BPC domain. Future work should compare pseudo-labeling with supervised models to continued pretraining with self-supervised  models to clarify the strengths and weaknesses of each approach.
\section{Acknowledgements}
We are grateful to Karen Livescu and Chris Graziul for their assistance with the Chicago BPC dataset, and we thank them and  Robert Cole Molloy for discussing various aspects of this work.

\newpage
\bibliographystyle{IEEEtran}
\bibliography{references}

\clearpage
\onecolumn
\begin{appendices}
\section{LLM-as-a-judge: Standard Prompt}

\label{app:standard_prompt}

Below is the standard prompt we use for LLM-as-a-judge filtering on Baltimore pseudo-labels.

\begin{tcolorbox}[
    colback=gray!5,          % Light gray background for the body
    colframe=gray!60,        % Medium gray border line
    arc=4mm,                 % Smooth rounded corners matching your layout
    boxrule=0.6mm,           % Crisper border definition
    left=15pt, right=15pt, top=14pt, bottom=14pt, % Spacious interior margins
    before upper={\setlength{\parindent}{0pt}},   % Keeps paragraphs flush left
    before lower={\setlength{\parindent}{0pt}},   % Keeps user segment flush left
    breakable                % Safely flows across page breaks if expanded
]
% --- SYSTEM PROMPT HEADER ---
{\large\bfseries System Prompt:}\\[0.5em]
You are an expert evaluator of automatic speech recognition (ASR) transcripts.
The audio being transcribed is police radio communication from the city of Baltimore, Maryland. The audio contains communication of dispatch assignments in response to calls for service from the 911/311 system in Baltimore, but they also capture police activity that emerges in response to events on the ground that require some degree of notification of officers working in the same police district and/or coordination among officers assigned to the same sector or post. The audio is recorded over radio channels and may contain background noise, police-specific vocabulary such as 10- codes, Baltimore street names and locations, call signs and dispatcher speech. 

\vspace{0.8em}
Your task is to judge whether a given ASR transcript is correct or not.
Consider whether the transcript looks like plausible speech from this domain — coherent words,
expected terminology, etc.
\vspace{0.8em}
Output ONLY the digit 1 or 0:
  1 = the transcript appears correct / reasonable
  0 = the transcript appears incorrect, garbled, or implausible
No explanation, no other text — just 1 or 0.

\vspace{0.8em}

% --- USER PROMPT HEADER ---
{\large\bfseries User Prompt:}\\[0.5em]
Transcript:
\end{tcolorbox}
\newpage
\section{LLM-as-judge: Rule-Based Prompt}
\label{app:rule-based_prompt}
Below is the rule-based prompt we use with Meta-Llama-3-8B-Instruct. 
\begin{tcolorbox}[
    colback=gray!5,          % Light gray background for the body
    colframe=gray!60,        % Medium gray border line
    arc=4mm,                 % Smooth rounded corners matching your layout
    boxrule=0.6mm,           % Crisper border definition
    left=15pt, right=15pt, top=14pt, bottom=14pt, % Spacious interior margins
    before upper={\setlength{\parindent}{0pt}},   % Keeps paragraphs flush left
    before lower={\setlength{\parindent}{0pt}},   % Keeps user segment flush left
    breakable                % Safely flows across page breaks if expanded
]
% --- SYSTEM PROMPT HEADER ---
{\large\bfseries System Prompt:}\\[0.5em]
You are an exceptionally strict, zero-tolerance quality control evaluator for police radio Automatic Speech Recognition (ASR) transcripts. Your goal is to isolate only the most flawless, high-fidelity transcripts for model training.
\vspace{0.8em}
The audio context is: police radio communication from the city of Baltimore, Maryland. The audio contains communication of dispatch assignments in response to calls for service from the 911/311 system in Baltimore, but they also capture police activity that emerges in response to events on the ground that require some degree of notification of officers working in the same police district and/or coordination among officers assigned to the same sector or post. The audio is recorded over radio channels and may contain background noise, police-specific vocabulary such as 10- codes, Baltimore street names and locations, call signs and dispatcher speech. 
\vspace{0.8em}

CRITICAL INSTRUCTION ON FORMATTING: Do NOT reject a transcript for minor numerical formatting variations (e.g., "10-4" vs "ten four", or "1014" vs "ten fourteen"). A downstream normalizer handles these, so they count as CORRECT. 
\vspace{0.8em}

Instead, you must ruthlessly judge the SEMANTIC and OPERATIONAL validity. You will also be given the duration of the audio file associated with the transcript. Reject the transcript immediately and output 0 if it contains ANY of the following flaws:
\vspace{0.8em}

1. CONTEXTUAL NONSENSE: The words technically form a grammatical English phrase but make absolutely zero sense in a law enforcement or dispatch context (e.g., phrases like "life coach records", "hard escape", "harper rope", "Adam Tutu", "riding an increase", or "put that through my blog"). \\
2. MISMATCH BETWEEN AUDIO DURATION AND TRANSCRIPT: The audio duration is very short (1-2 seconds) and yet the transcript is several words long, or vice versa.  \\
3. CASUAL OR COLLOQUIAL HALLUCINATIONS: The transcript includes non-professional, casual filler words that are rare in formal dispatch (e.g., "Y'all", "man", "with all God's permission").\\
4. PHONETIC CODE MANGLES: Standard 10-codes or alphanumeric unit callsigns are corrupted into phonetically similar but wrong words (e.g. "Charlie for once" instead of "Charlie 41").
\vspace{0.8em}
ONLY output a 1 if the transcript is a high-fidelity, pristine, and clean representation of standard professional police radio chatter. If you have even any doubt whatsoever that a phrase sounds corrupted or contextually bizarre, mark it 0.
\vspace{0.8em}
Output ONLY the digit 1 or 0:
 1 = The transcript is operationally flawless and contextually correct.
 0 = The transcript has a contextual error, truncation, hallucination, or flaw.
\vspace{0.8em}
No explanation, no other text — just 1 or 0.
\vspace{0.8em}

% --- USER PROMPT HEADER ---
{\large\bfseries User Prompt:}\\[0.5em]
Audio Duration: \\

Transcript:
\end{tcolorbox}
\newpage
\section{Hyperparameters for Model Training}
We finetune Whisper large-v3  using a single NVIDIA A100 GPU for a maximum of 100 epochs with a batch size of 8 with a learning rate of $5 \times 10^{-5}$, and the fused AdamW optimizer. We use bf16 precision and gradient norm clipping for numerical stability. We perform validation every 100 steps, triggering early stopping when performance plateaus for more than 10 steps. 
\\

We finetune the Qwen3-ASR 1.7B model for up to 10 epochs on a single NVIDIA A100 GPU using a batch size of 4 and a learning rate of $2 \times 10^{-5}$. To prevent overfitting, we implement early stopping with a patience of 10. We use the same hyperparameters to finetune both models throughout this work.
\label{app:model_training}
\newpage

\section{Un-normalized WERs}
\label{app:un-normalized}

We report the unnormalized WERs for the experiments in Table \ref{tab:horizontal_main_results}.

\begin{table}[htbp]
\centering
% \begin{table}[htbp] % Changed to single-column width since it's now a compact 3 columns
\caption{Different pseudo-label filtering results on Qwen3-ASR}
\label{tab:Qwen3-ASR_unnorm}

\begin{tabular}{lcc}
\toprule
& \textbf{Baltimore} & \textbf{Chicago} \\
\cmidrule(lr){2-2}\cmidrule(lr){3-3}
\textbf{Method} & \textbf{\makecell{Unnormalized\\Test Set WER}} & \textbf{\makecell{Unnormalized\\Test Set WER}} \\ 
\midrule
OTS              & 0.6749 & 0.3761 \\ \hline
Oracle-filtering & 0.6731 & 0.3240 \\ \hline
Unfiltered       & 0.6799 & 0.3646 \\ 
Log-prob         & 0.6776 & 0.3682 \\ 
STAR             & 0.6864 & 0.3738 \\ 
LLM-as-judge     & 0.6311 & 0.3561 \\ 
\bottomrule
\end{tabular}
% \end{table}
\vspace{1.5em}
% \begin{table} % Changed to single-column width since it's now a compact 3 columns
\caption{Different pseudo-label filtering results on Whisper}
\label{tab:whisper_unnorm}
\centering
\begin{tabular}{lcc}
\toprule
& \textbf{Baltimore} & \textbf{Chicago} \\
\cmidrule(lr){2-2}\cmidrule(lr){3-3}
\textbf{Method} & \textbf{\makecell{Unnormalized\\Test Set WER}} & \textbf{\makecell{Unnormalized\\Test Set WER}} \\ 
\midrule
OTS              & 0.6010 & 0.7292 \\ \hline
Oracle-filtering & 0.5963 & 0.4374 \\ \hline
Unfiltered       & 0.6304 & 0.5403 \\ 
Log-prob         & 0.6078 & 0.5029 \\ 
STAR             & 0.5983 & 0.4858 \\ 
LLM-as-judge     & 0.6217 & 0.5230 \\ 
\bottomrule
\end{tabular}
\end{table}
\newpage

\section{Cross-City Finetuning}
Table \ref{tab:cross-city-ft} shows the results of cross-city finetuning. Whisper is finetuned with manual transcripts from one city's BPC train set and then conducts inference on the other city's test set. Results are normalized with the Whisper normalizer.  We show normalized OTS performance on each city's test set for comparison. 

\begin{table}[htbp]
\caption{Cross-city Finetuning}
\label{tab:cross-city-ft}
\centering
\begin{tabular}{llll}
\toprule
\textbf{Base Model} & \textbf{Train Set} & \textbf{Test Set} & \textbf{\makecell{Test Set\\WER}} \\
\midrule
Whisper & - & Chicago & 1.1104 \\
Whisper & Baltimore & Chicago & 1.1867 \\ \hline
Whisper & - & Baltimore & 0.8914 \\
Whisper & Chicago & Baltimore & 0.8631 \\
\bottomrule
\end{tabular}
\end{table}
\end{appendices}
\end{document}